\documentclass[runningheads]{llncs}
\usepackage{graphicx}
\begin{document}
\title{Of Mice and Pose: 2D Mouse Pose Estimation from Unlabelled Data and Synthetic Prior}

%
%
%
\titlerunning{Of Mice and Pose}
%
\author{
Jose Sosa \inst{1} \and
Sharn Perry\inst{2} \and
Jane Alty \inst{2,3} \and
David Hogg \inst{1}
}
\authorrunning{J. Sosa et al.}
%
\institute{School of Computing, University of Leeds, United Kingdom.\and
Wicking Dementia Research and Education Centre, College of Health and Medicine,
University of Tasmania, Australia. \and
School of Medicine, College of Health and Medicine, University of Tasmania, Australia.\\
\email{\{scjasm,d.c.hogg\}@leeds.ac.uk}, \email{\{sharn.perry, jane.alty\}@utas.edu.au}}
\maketitle              
\begin{abstract}
 Numerous fields, such as ecology, biology, and neuroscience, use animal recordings to track and measure animal behaviour. Over time, a significant volume of such data has been produced, but some computer vision techniques cannot explore it due to the lack of annotations. To address this, we propose an approach for estimating 2D mouse body pose from unlabelled images using a synthetically generated empirical pose prior. Our proposal is based on a recent self-supervised method for estimating 2D human pose that uses single images and a set of unpaired typical 2D poses within a GAN framework. We adapt this method to the limb structure of the mouse and generate the empirical prior of 2D poses from a synthetic 3D mouse model, thereby avoiding manual annotation. In experiments on a new mouse video dataset, we evaluate the performance of the approach by comparing pose predictions to a manually obtained ground truth. We also compare predictions with those from a supervised state-of-the-art method for animal pose estimation. The latter evaluation indicates promising results despite the lack of paired training data. Finally, qualitative results using a dataset of horse images show the potential of the setting to adapt to other animal species.
 
 
\end{abstract}
\section{Introduction}
The study of neurodegenerative human diseases, such as Alzheimer's disease \cite{heuer2012nonhuman,saito2014single}, Parkinson's disease \cite{lee2012animal}, and Amyotrophic Lateral Sclerosis (ALS) \cite{philips2015rodent}, usually involves using animal models. Mice are the preferred and most extensively utilised animals for such studies because of their genomics similarity with humans and the accumulated knowledge on manipulating their DNA \cite{fisher2019mouse}. Due to this tight relationship between mice and the ongoing research on neurodegenerative human diseases, developing tools to observe, describe, and measure mouse behaviour has become crucial \cite{mathis2020deep}.

Some years ago, prior to the adoption of computer vision techniques, making such measurements meant tons of manual labour \cite{noldus2000observer,olivo1988monitoring}. For example, if someone wanted to measure the position of the mouse's limbs. It implies recording the animal, looking at each video frame, and manually identifying each required body part. Then, it is evident that manual inspections on large videos can be time-consuming and lead to observation errors. Early computational approaches attempt to minimise human intervention in analysing animal recordings. Some tools involve placing physical markers on the animal's body or require painting the body parts to track \cite{bender2010computer,spink2001ethovision}. Apparent limitations of these techniques are that the physical markers can interfere with the animal's behaviour, and the information that can be extracted is inherently limited by the positioning of the markers or the painted areas. Other approaches use sophisticated and expensive equipment to acquire particular images, which results in costly experiments and problems for deployment and replication \cite{card2008visually,kirkpatrick1991computerized,gershow2012controlling}.

Newer computer vision tools for tracking body parts of animals \footnote{https://mousespecifics.com/digigait/}\footnote{ https://www.noldus.com/catwalk-xt} become less dependent on physical markers, i.e. markerless. Unfortunately, these tools still needed considerable human intervention for pre-processing and post-processing video data. Supervised deep learning approaches have recently become state-of-the-art for pose estimation and tracking of humans and animals \cite{mathis2018deeplabcut,pereira2019fast,graving2019deepposekit}. Performance of these techniques often depends on the amount and variability of annotated data for training, which is hard to obtain for some animal species. Thus, there remains an urgent need to develop methods for tracking animal pose that require minimal human effort in training for a new animal domain and operational use. This can be achieved by reducing the need for manual pose annotation of images.

In this paper we tackle the challenging task of predicting 2D mouse poses from unlabelled images. Different from previous deep learning approaches that generally rely on fully supervised frameworks, we adopt a self-supervised 2D pose estimator from the human domain \cite{jakab:2020}. This method utilises a GAN architecture to learn 2D human poses. During training, it assumes the availability of unlabelled images and an unpaired prior of 2D pose annotations, generally from the same dataset. Our proposal relaxes much more the assumptions about data by building the needed prior of 2D poses using data generated from a 3D model of a generic mouse \cite{bolanos2021three}. Evidently, incorporating synthetic data also provides more flexibility to train the model with entirely unlabelled datasets, which is common for many animal recordings outside of computer vision. Furthermore, our method shows promising results in generating 2D poses for other types of animals, e.g. horses. This demonstrates the viability of our approach to be rapidly deployed to different domains without the burden of annotating data.

\section{Related Work}\label{sec:related-work}
\subsection{Deep Learning Methods for Animal Pose Estimation}
Analogous to the definition of human pose estimation \cite{liu2015survey}, animal pose estimation refers to the task of estimating the geometrical configuration of body parts of an animal. This problem has gained increasing attention because of research applications in many different disciplines, including Biology, Zoology, Ecology, Biomechanics \cite{sheppard2022stride} and Neuroscience \cite{mathis2020deep}. Compared with human pose estimation, it is still relatively under-explored, principally due to the variability of animal species, and the need for species-specific labelled datasets. Nevertheless, a lot of effort has gone into developing and adapting deep learning models to estimate 2D and 3D animal pose, exploiting similarities between many species of animal. For example, monkeys \cite{yao2019monet,negrete2021multiple,bala2020automated} share similar skeletal structure with humans. Large quadrupeds, such as farm animals \cite{chen2021behaviour,mathis2021pretraining,russello2022t,riekert2020automatically} and dogs \cite{biggs2020left,shooter2021sydog,kearney2020rgbd} present similarities between their skeletal forms.

Automatic 2D pose estimation has also been applied successfully on smaller animal species such as mice. As with larger animals, deep learning methods for pose estimation have been based mostly on supervised methods developed for human pose estimation. Their performance is therefore limited by the availability and correctness of annotated data. For example, DeepLabCut (DLC) \cite{mathis2018deeplabcut} adapts a pretrained ResNet with deconvolutional layers \cite{insafutdinov2016deepercut} to estimates the 2D pose of small animals under laboratory conditions, such as mice and flies. LEAP \cite{pereira2019fast} also uses an earlier model from the human pose estimation domain \cite{tompson2014joint} to solve the same task. DeepPoseKit \cite{graving2019deepposekit} employs a similar method to estimate 2D animal pose. It uses a network architecture that improves the processing speed based on fully convolutional densenets \cite{huang2017densely,jegou2017one} and stacked hourglass modules \cite{newell2016stacked}. More recently, OptiFlex \cite{liu2021optiflex} exploits the temporal information in video data by incorporating flowing convnets \cite{pfister2015flowing} into their network architecture. They report similar performance to previous methods \cite{mathis2018deeplabcut,pereira2019fast,graving2019deepposekit} on estimating the pose of small animals, e.g. mice, fruit flies, and zebrafish. 

Perhaps the most popular of these approaches is DeepLabCut. Many subsequent methods adopt it to estimate not only mouse pose, but also pose for a wide variety of other animal species \cite{sato2021markerless,karashchuk2021anipose,raman2022markerless,labuguen2019primate,wang2021identifying,liu2020video,farahnakian2021multi}. A common feature of DeepLabCut, DeepPoseKit, LEAP, and OptiFlex is their reliance on manual annotation of pose in multiple video frames. Even though they normally provide a Graphical User Interface (GUI) for doing the annotation, the process is still time consuming, error prone, and requires specialised knowledge to infer pose correctly. Futhermore, the number of frames to annotate for good generalisation is hard to predict and therefore ultimately determined empirically. In contrast, through adapting a recent self-supervised approach from the human domain, we completely remove the need for manual annotation, making training and testing more straightforward.

\subsection{Animal Pose Estimation with Synthetic Data}
One alternative to avoid manual annotation for training deep learning methods for animal pose estimation is the use of synthetic data. Using an artificial animal model allows producing many synthetic images and their corresponding annotations with less time and effort than manually annotating actual data \cite{bolanos2021three}. In this context, Mu et al. \cite{mu2020learning} proposes a semi-supervised pose-estimation framework trained in a supervised fashion using synthetically rendered images and ground truth pose annotations from 3D Computer-Aided Design (CAD) models. Then, they perform self-supervised domain adaption with a small portion of actual data to minimise the domain gap. They successfully estimate 2D poses for large animals with similar skeletal structures, such as tigers, horses, and dogs. Some other works relying on synthetic data also focus on the domain adaptation process after learning the animal pose with synthetic data under supervised paradigms \cite{li2021synthetic,jiang2022prior}. We adopt a related approach to \cite{mu2020learning} by using an existing 3D geometric mouse model \cite{bolanos2021three}, except that we do not use rendered images as in supervised settings. We only utilise the synthetically generated 2D poses as a prior for training the method. In particular, we use this prior on 2D poses within a GAN framework that allows our whole model to learn poses not necessarily appearing in the prior, eliminating the need for domain adaptation as in \cite{mu2020learning,li2021synthetic,jiang2022prior}.

Synthetic data also plays a significant role in learning more complex forms of 3D animal poses. For instance, \cite{zuffi20173d} inspired by the success of human shape models, SMPL \cite{loper2015smpl} generates data from toy figurines of animals to learn statistical shape models (SMAL). Later, \cite{zuffi2018lions} propose SMALR, which is an extension of the previous SMAL model. It introduces a regularisation for the deformation of the animal shape to make it appear more detailed and realistic. Subsequent work has \cite{biggs2020left,ruegg2023bite,zuffi2019three} adapted the SMAL model to work with particular animal species like dogs and zebras. In contrast to learning to fit 3D shape models from 3D scans, other approaches explore the possibility of learning 3D animal models from less complex representations, like multi-view 2D images, or user-clicked 2D images \cite{cashman2012shape,goel2020shape,kanazawa2016learning}. However, the final shape representation of those models is less realistic and detailed than those produced using SMAL or SMALR. These methods have produced 3D shape models for various animal species, typically focused on large quadrupeds like tigers, dogs, and zebras. Unfortunately, creating sophisticated models for all animal species is still impractical.

Bolaños \cite{bolanos2021three} has taken inspiration from previous synthetic models of large animals to develop a similar model for mice. This 3D CAD model simulates semi-random behavioural patterns from real mice and incorporates the 3D structure of bones and joints. The model has successfully created training data for famous supervised 2D and 3D mouse pose estimation approaches \cite{mathis2018deeplabcut,nath2019using}. Nevertheless, there is still an unexplored opportunity to utilise the same model to generate data for training pose estimation models with lower levels of supervision. We demonstrate this by relying on a recent self-supervised method that learns to estimate 2D human poses solely from unlabelled images and a prior on unpaired 2D poses. We follow the same idea, but instead of taking the unpaired pose annotations from the dataset to build the prior, we generated them with a 3D mouse model \cite{bolanos2021three}. Note that we do not utilise paired synthetic images and pose annotations like in previous works \cite{mu2020learning,li2021synthetic,jiang2022prior,bolanos2021three}, we discard the synthetic images and only use synthetic 2D poses. This means that our model is trained using actual unlabelled images and a smaller set of artificially generated 2D poses.

\section{Method}\label{sec:method-comp}

\begin{figure}[ht]
\centering
\includegraphics[width=\textwidth]{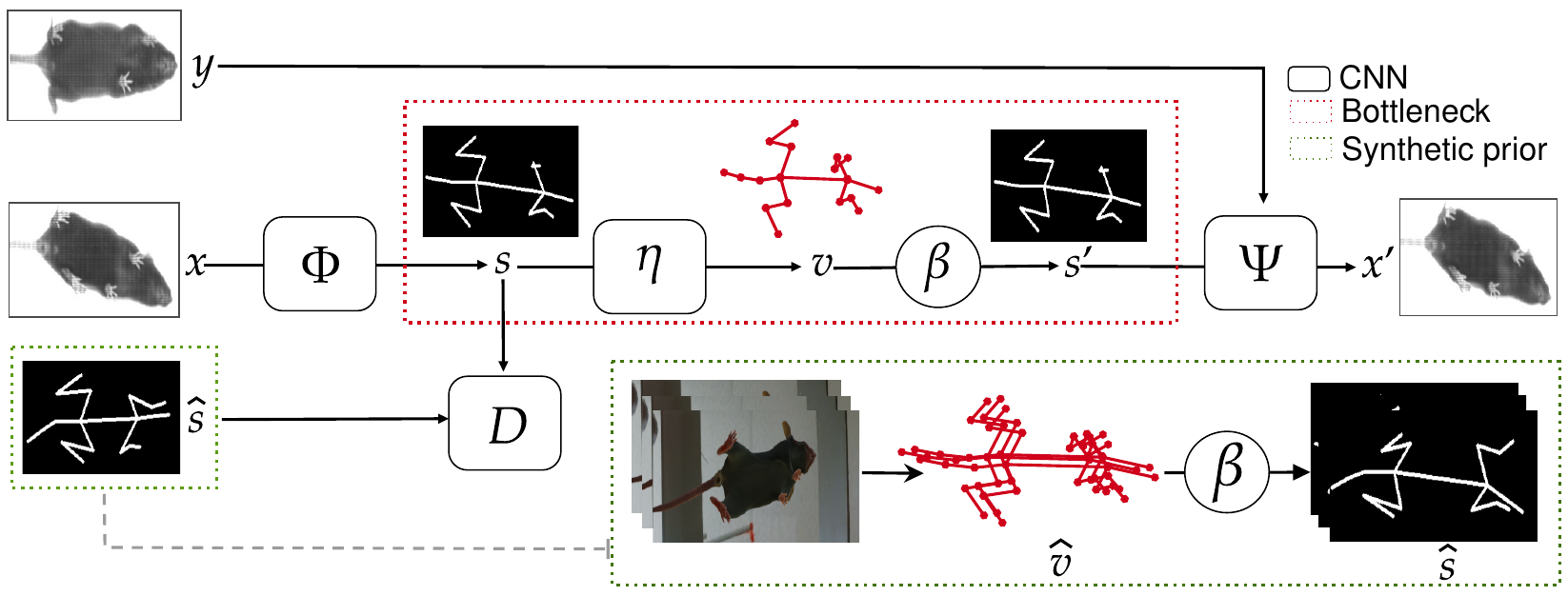}
\caption{2D pose estimator. We use a self-supervised 2D pose estimator from the human domain, which we adapt to work with mice. Differently to the original implementation, we build prior of 2D poses using synthetic data from a 3D model of a generic mouse.}
\label{pose-extractior}
\end{figure}

Our method produces a mapping from full body images to the 2D pose of a mouse, as shown in Fig.\ref{pose-extractior}. The pose is represented as an articulated tree structure of 2D line segments corresponding to the parts of the body such as snout, tail, hind limbs, and forelimbs. The method extends the self-supervised approach of \cite{jakab:2020}, which estimates human 2D pose. This 2D pose estimator learns from unlabelled images and uses a set of unpaired 2D poses as an empirical prior. This removes any dependence on paired annotated data. However, the method requires a set of manual 2D pose annotations for a subset of images from the dataset, albeit the pairing is discarded. We adapt this approach by changing the pose topology to a mouse model. We also generate an empirical prior for 2D mouse pose by projecting from an existing 3D mouse model, which removes the need for manual pose annotation altogether.

The pose-estimator is obtained by training a conditional auto-encoder to map from an image $x$, depicting a mouse, to a reconstructed image $x'$ that is as similar as possible. The synthesis of the output image is conditioned on an auxiliary mouse image $y$ depicting a fixed pose. The auto-encoder has a bottleneck that encodes the 2D pose as a set of joint positions $v$. Once trained, our pose predictor is the initial encoder from this network, which maps from an input image to a 2D pose. This mapping is in two steps, consisting of a Convolutional Neural Network (CNN) $\Phi$ mapping from the image $x$ to a skeleton image $s$, followed by a second CNN $\eta$ mapping from the skeleton image $s$ to the 2D pose $v$. The decoder mapping from the 2D pose $v$ to the output image $x'$ is also in two steps, consisting of a differentiable function $\beta$ which maps the 2D pose $v$ to a skeleton image $s'$; and a CNN $\Psi$ mapping from the skeleton image $s'$ to the output $x'$. The second mapping $\Psi$ takes an auxiliary image $y$ as an additional input to compensate for the missing appearance information in $s'$.

We train the model with a dataset of images $\{x_1 \cdots x_N\}$, depicting mice in different poses, and our empirical prior of 2D poses. We use a similar loss function as in \cite{jakab:2020}, which contains three terms. The first penalises the difference between the generated image $x'$ and the input $x$ via a perceptual loss. The second term is a regression loss to evaluate the mapping from skeleton image $s$ to the 2D joint positions in $v$. The third term is an adversarial loss to assess the authenticity of the skeleton images generated in the encoder. In the following sections we provide details on the components of the model, the empirical prior, loss function, and training.

The whole pipeline for the conditional auto-encoder is as follows:

\begin{equation} \label{eq:2}
    x=\Psi(\beta(v)\circ\eta(s)\circ\Phi(x), y)
\end{equation}

We can see the mapping as an autoencoder from input image $x$ to output image $x'$ in which the 2D pose $v$ emerges as an intermediate representation. In training the network, a perceptual loss \cite{johnson2016perceptual} compares each input image $x$ with the  reconstructed image $x'$:

\begin{equation}\label{eq:perc}
\mathcal{L}_{perc} = \frac{1}{N}\sum_{i=1}^N \Vert \Gamma(x'_i) - \Gamma(x_i)\Vert_2^2
\end{equation}

where $\Gamma$ is a pre-trained VGG network \cite{simonyan2014very} with the classification stage removed to utilise the final feature encoding.

A CNN serves as the discriminator network $D$, which outputs a probability that an input skeleton image comes from the prior distribution of skeleton images. Thus, $D$ measures the extent to which a skeleton image $s$ looks like an authentic skeleton image from the empirical prior distribution. Note that contrary to \cite{jakab:2020}, our prior $\{\hat{v}_{j}\}^M_{j=1}$ is synthesised by projecting from a 3D mouse model and does not require manual annotation of poses. We obtain the skeleton images $\{\hat{s}_j\}_{j=1}^{j=M}$ via $\beta$, i.e.
$\{\hat{s}_j = \beta({\hat{v}_{j}})\}^M_{j=1} $, then we compare this distribution $p_{data}(\hat{s})$ with the distribution $p_{data}(s)$ from the predicted skeleton images  $\{s_i = \Phi({x_{i}})\}^N_{i=1} $ by means of the adversarial loss \cite{mao2017least}:

\begin{equation} \label{eq:disc}
    \mathcal{L}_D = \frac{1}{M} \sum_{j=1}^M D({\hat{s}}_j)^2 + \frac{1}{N}\sum_{i=1}^N(1 - D(s_i))^2
\end{equation}

Finally, we derive a loss from $\eta$ and $\beta$, which combines the 2 terms as follows:

\begin{equation}
\label{eq:reg}
    \mathcal{L}_{\eta} = \Vert \eta(\hat{s}) - \hat{v} \Vert^2 + \lambda \Vert \beta(\eta(s)) - s \Vert^2 
\end{equation}

The first term uses unpaired 2D poses from the prior, while the second one utilises the pose on the predicted skeleton image $s$. The last term ensures that the network learns poses that appear on the training images but not necessarily on the prior. The balancing coefficient $\lambda$ is set to $0.1$ in our experiments.

\subsubsection{2D synthetic prior.}  
\label{sub-sec:sythetic-prior} We entirely generate the 2D pose prior required for the discriminator $D$ using synthetic data. In particular, we adopt a synthetic 3D model of a mouse \cite{bolanos2021three}. This animated mouse model simulates synthetic behavioural data using animation and semi-random joint movements. We keep the original joint-constrained movements of the freely moving mouse model. We animate and render \footnote{ We use Blender to make the videos and extract the 2D poses from the mouse model.} the different scenes with the synthetic model and extract the 2D coordinates of 18 joints on the body of the mouse: Snout, Vertebral column base and end (VB and VE), three points located along the tail (TB, TM, and TE), left/right elbows (LE and RE), left/right knees (LK and RK), and two points (tip and top) for each left/right fore and hind limbs (LFP$^{-/+}$, RFP$^{-/+}$, LHP$^{-/+}$, RHP$^{-/+}$). Note that this notation will be used through the paper. Finally, we use those joint positions to create their respective skeleton image, as shown in Fig.\ref{pose-extractior}. Overall, our prior consists of 15,408 different 2D poses transformed into skeleton images.

\subsubsection{Training.} Following \cite{jakab:2020} we use a perceptual loss  $\mathcal{L}_{perc}$ (\ref{eq:perc}), an adversarial loss $\mathcal{L}_{D}$ (\ref{eq:disc}), and a regression loss (\ref{eq:reg}) in training the convolutional networks $\Phi$, $\eta$, and $\Psi$. Note that $\beta$ is not a learnable function. The overall loss $\mathcal{L}$ is given by:

\begin{equation}\label{eq:3}
    \mathcal{L} = \mathcal{L}_{D} + \mathcal{L}_{\eta} + \mathcal{L}_{perc}
\end{equation}

We train the pose estimator using unlabelled images. In particular, each batch is formed by randomly sampling images $(x, y)$, and a random sample $\hat{v}$ from the synthetic 2D poses, which is then transformed to skeleton image $\hat{s}$. The input images $x$ and $y$ were resized to $128 \times 128$ pixels. We set the batch size to 32 and use Adam optimiser \cite{kingma2014adam} with a learning rate of $2\times10^{-4}$, $\beta_1=0.5$, and $\beta_2=0.999$. Unlike \cite{jakab:2020} who use a pretrained $\eta$, we train all the neural networks $\Phi$, $D$, $\eta$, and $\Psi$ from scratch by optimising the loss function in Equation \ref{eq:3}. During testing, we only rely on the trained networks $\Phi$ and $\eta$, to map from an input image to a 2D pose. Specifically, we input the image $x$ through $\Phi(x)$ to obtain the skeleton image $s$, and then use this with $\eta(s)$ to get the final 2D pose $v$.

\section{Experiments} \label{sec:exp}
\subsubsection{Dataset.}
Our dataset contains images from $40$ videos of rodent models of ALS of different genotypes\footnote{All the mice appearing in the recordings were bred and maintained at the Univeristy of Tasmania.}. Each video has around $13,120$ frames/images, with an original size of $658 \times 190$ pixels. We use half of the available videos to get the training images, and reserve the other half for evaluation purposes.

\textbf{Acquisition details.} The recordings were made using the Digigait\textsuperscript{TM} apparatus, which consists of a transparent treadmill and a camera placed underneath. Mice  at both 4 and 16 weeks of age were first acclimatised in the apparatus and then encouraged to run on the treadmill at $10 cm/s$, $20 cm/s$ and $30 cm/s$ for a minimum of 10 seconds. The camera captures the mice on video as they move on the treadmill. Mice were gently encouraged to run by taps to their rear by the experimenter if needed. At the end of the trial, the mice were returned to their home cage. The average duration of each video is $80$ seconds, i.e. most of the mice ran for at least 20 seconds at each speed with 10 seconds transitions between speeds without running.


\subsubsection{Results.}\label{sec:results}
\begin{figure}
\centering
\includegraphics[width=\textwidth]{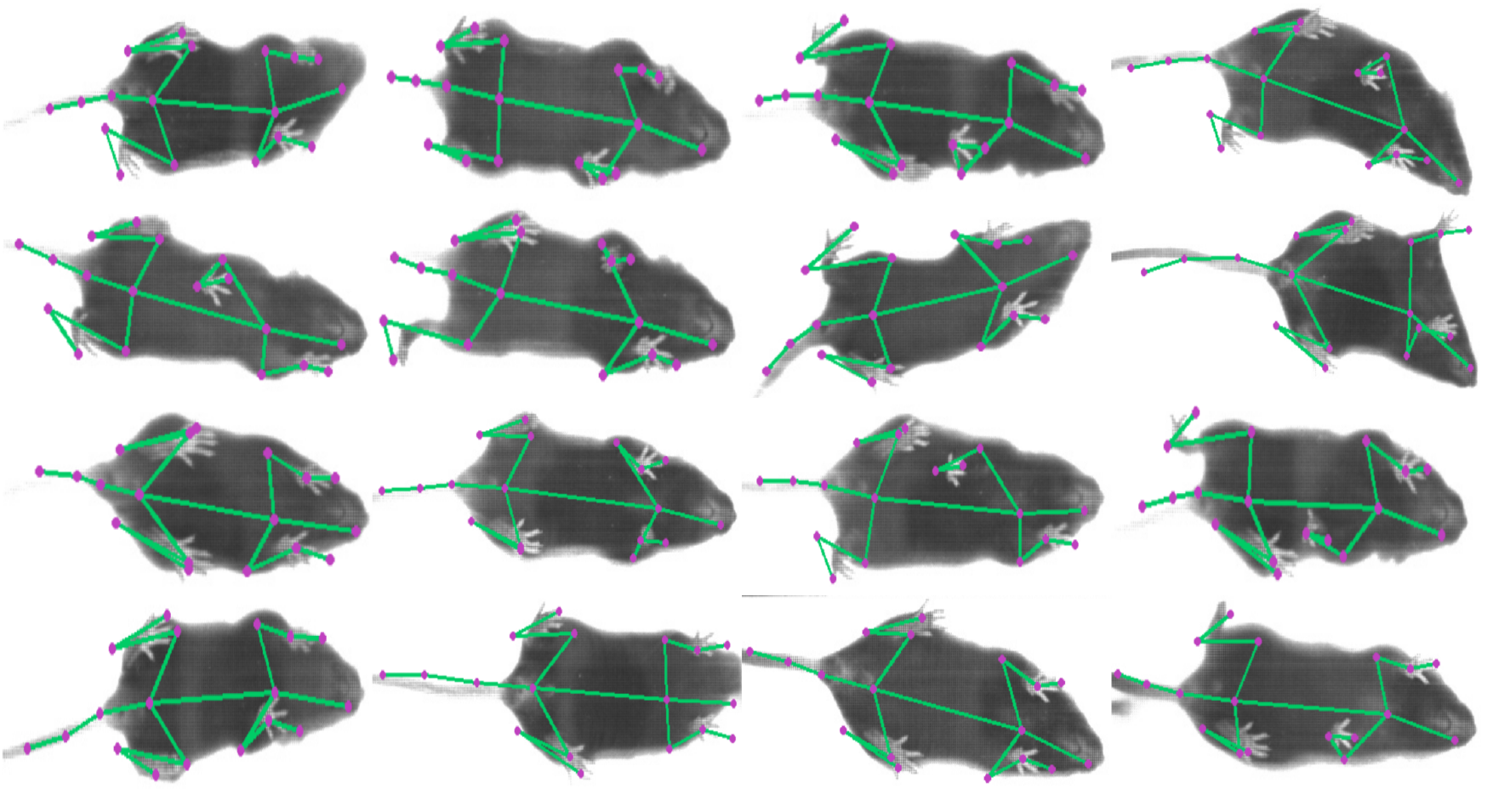}
\caption{Estimated 2D poses using our method. During training we use real images and the synthetic pose prior: \textbf{RI + SP}.}
\label{2d-pose-qali-results}
\end{figure}

Given an unlabelled image depicting a mouse, our trained model produces a 2D representation of the mouse pose composed of 18 joint positions. Fig.\ref{2d-pose-qali-results} shows some of those predicted 2D poses. Since our dataset does not contain annotations for the joint positions, we manually annotated 2D poses for some images in the test videos to provide ground truth for a quantitative measure of prediction performance. We compare pose predictions with ground-truth on this test set using the Mean Per Joint Position Error (MPJPE). The first row of Table \ref{tab:mpjpe} shows the MPJPE in pixels between the predicted positions for each of the joints composing the mouse pose and their respective ground truth annotations. MPJPE is reported w.r.t to the original image dimensions: $658 \times 190$ pixels.

\begin{figure}[ht]
\centering
\includegraphics[width=\textwidth]{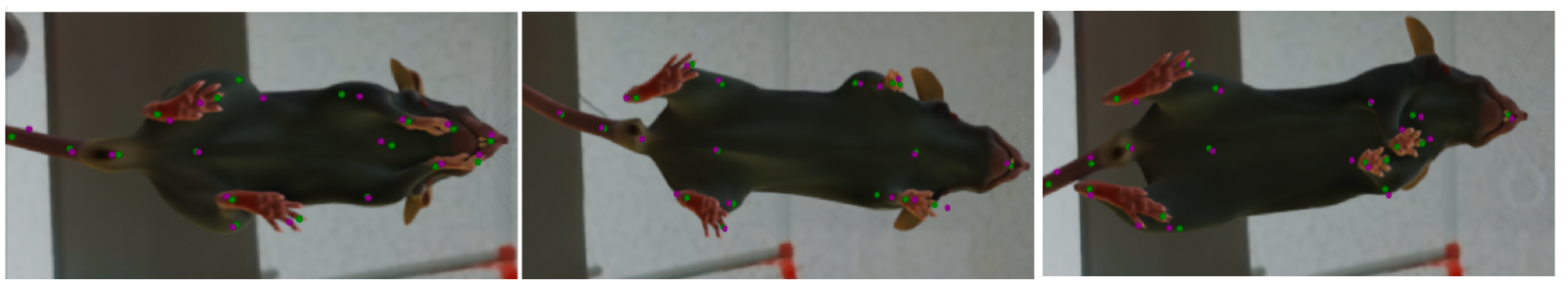}
\caption{Predicted 2D poses using the model trained on synthetic images and synthetic prior \textbf{(SI + SP)}. Images rendered from synthetic model showing their respective predicted (purple dots) and ground truth 2D poses (green dots).}
\label{pose-estimation-syn}
\end{figure}

In addition to the previous experiment, we train and evaluate our model using synthetic images and synthetic unpaired poses (SI + SP). Note that the synthetic 2D poses on the prior are not annotations of the training images. We train the model with different sequences of images synthetically generated from the 3D mouse model and test it using a different set of synthetic images. We use the 2D ground truth annotations for 18 joint positions extracted from the mouse model and compare them with the predicted poses. We report the MPJPE for each joint position in the second row of Table \ref{tab:mpjpe} and a few visualisations of results in Fig.\ref{pose-estimation-syn}.

\begin{table}[h]
    \caption{MPJPE of predicted poses. \textbf{RI + SP} denotes use of the method trained with \textbf{R}eal \textbf{I}mages and \textbf{S}ynthetic \textbf{P}rior. \textbf{SI + SP} denotes use of the method trained with \textbf{S}ynthetic \textbf{I}mages and \textbf{S}ynthetic \textbf{P}rior.}
    \label{tab:mpjpe}
    \centering
    \begin{tabular}{|l|c|c|c|c|c|c|c|c|c|c|c|c|c|c|c|c|} 
    \hline
    \textbf{Joints} & Snout & VCB & VCE	& TB &	TM & TE	& RE &	RFP$^-$ &	RFP$^+$ & \textbf{Avg.} \\
     & LE	 &LFP$^-$	&LFP$^+$&	RK&	RHP$^-$ &	RHP$^+$&	LK&	LHP$^-$ &	LHP$^+$ &  \\
    \hline
    \hline
    \textbf{RI + SP} &	13.4	&8.0&	5.6&	15.2&	17.8&	31.8&	14.7&	15.8&	14.8 &\\
     & 12.7&	10.5&	14.2&	7.6	&21.1&	11.5&	14.9&	11.7&	11.9&	\textbf{14.1}\\
    \hline
    \textbf{SI + SP} &	5.9&	4.0&	3.0	&3.7	&4.3&	6.2&	5.9&	6.6&	7.1& \\
     &	5.9	&6.9&	7.0&	4.1&	5.2&	6.0&	4.0&	5.1&	5.0&	\textbf{5.3}\\
    \hline
    \end{tabular}   
\end{table}

\subsubsection{DeepLabCut comparison.}
In the absence of a more extensive set of annotated data for evaluating all our predictions, we also report on a quantitative comparison with the predictions from a state-of-the-art supervised method for animal pose estimation: DeepLabCut \cite{mathis2018deeplabcut}. The motivation for performing this comparison is to show that our self-supervised approach can work similarly to this supervised method, removing the requirement to annotate 2D poses for training. To build the training set for DLC, we select a subset of 100 consecutive images from one video and label 18 joint positions in each one. We then use these images and their labelled 2D poses to train a DLC model in a supervised fashion. We follow the official implementation of DLC \cite{mathis2018deeplabcut}. Using the trained DLC model, we then predict the pose for unseen videos. 

\begin{figure}[ht]
\centering
\includegraphics[width=\linewidth]{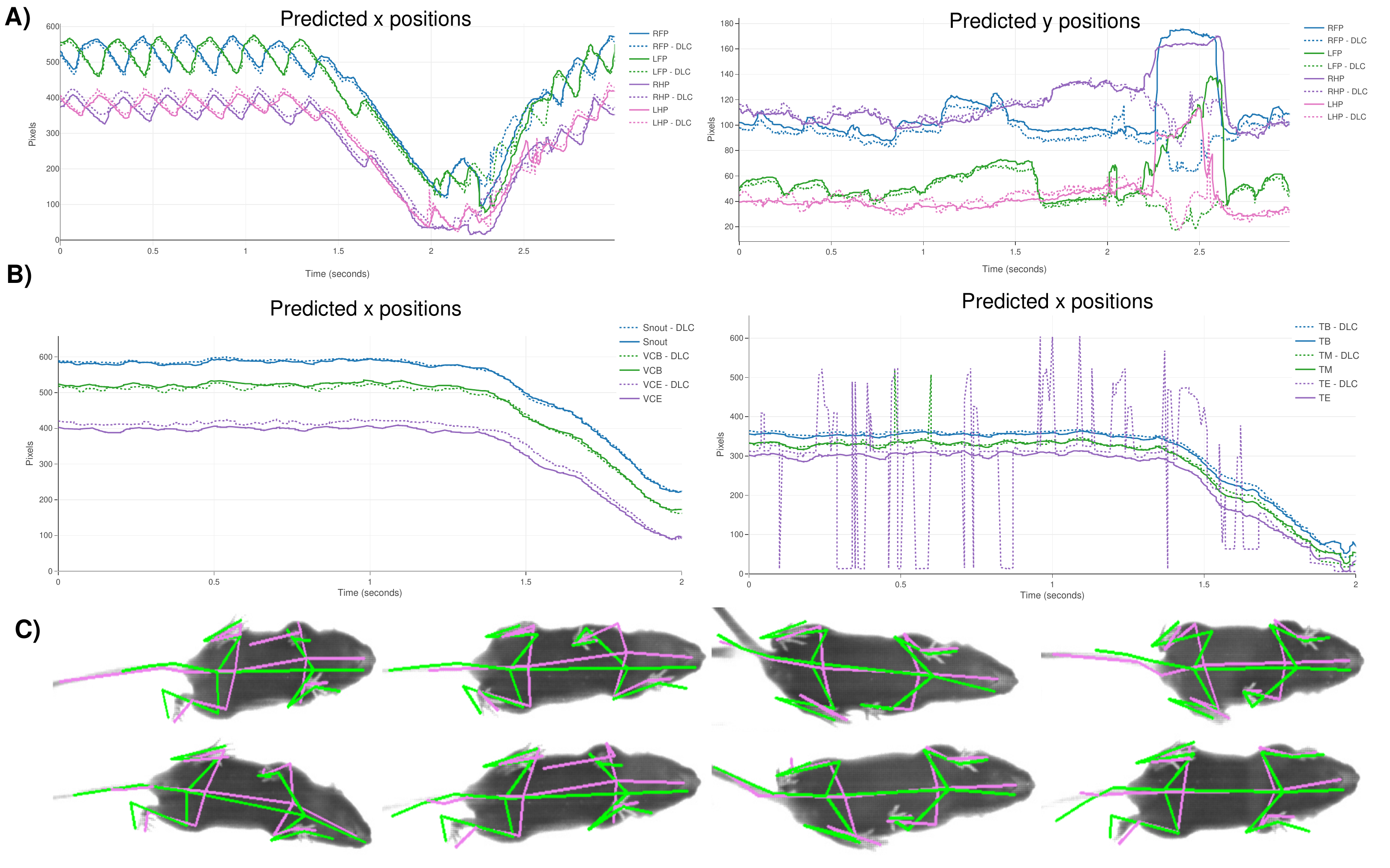}
\caption{Comparison of our predicted joint positions against the ones predicted by DLC.
\textbf{A)} Predictions for RFP, LFP, RHP, and LHP. 
\textbf{B)} Predictions for snout, vertebral column, and tail. 
\textbf{C)}Visual comparison of predicted poses by DLC and our method. Green - ours; pink - DLC.}
\label{all-results}
\end{figure}

We compare the predictions of our method against the ones produced by DLC. Each estimated body joint position is represented as a $(x,y)$ pair of coordinates on the image plane. Fig.\ref{all-results} summarises the results of our comparison. 
Note that each graph contains our estimated positions (indicated with lines) for a given joint together with the ones estimated by DLC (indicated by dotted lines). In the inset legend, we use the label `DLC' after the name of the joint to identify the predicted joint positions by DeepLabCut. The predictions of our method simply appear as the name of the joint. Finally, we assess quantitatively the predictions of DLC with the same ground-truth of that we used to evaluate predictions from the self-supervised method. Table \ref{tab:mpjpe-dlc} shows the MPJPE of DLC predictions and their respective ground truth. As expected, the overall MPJPE is lower for DLC. This may be explained in part by the use of supervision in training DLC, albeit on a limited dataset, and the consistency with which joint positions are manually located in producing ground-truth for the training and testing images.

\begin{table}
    \caption{MPJPE of predicted poses with DLC.}
    \label{tab:mpjpe-dlc}
    \centering
    \begin{tabular}{|l|c|c|c|c|c|c|c|c|c|c|c|c|c|c|c|c|} 
    \hline
    \textbf{Joints} & Snout & VCB & VCE	& TB &	TM & TE	& RE &	RFP$^-$ &	RFP$^+$ & \textbf{Avg.} \\
     & LE	 &LFP$^-$	&LFP$^+$&	RK&	RHP$^-$ &	RHP$^+$&	LK&	LHP$^-$ &	LHP$^+$ &  \\
    \hline
    \hline
    \textbf{DLC} &	4.7&	16.3&	18.2&	4.0&	7.2&	20.2&	7.8&	5.4	&5.1 &\\
     &6.5&	5.7&	9.8&	5.6&	5.3&	6.1&	7.2&	9.0&	8.3&	\textbf{8.5}\\
    \hline
    \end{tabular}
    
\end{table}

\subsubsection{Adaptation to other structures.} 
We demonstrate that our experimental setting, i.e. using synthetic prior and actual data for training, is adaptable to other animal structures. We build a dataset of horse images using a combination of individual frames from YouTube videos depicting horses in motion and horse images from the TigDog dataset \cite{del2015articulated}. The 2D poses for the synthetic prior come from the synthetic horse model of  \cite{mu2020learning}. We train the model using our dataset of approximately 30k horse images and a prior of 10k synthetic 2D poses. Once trained, we evaluate it utilising an out-distribution dataset \cite{borenstein2004combining} and some horse images from the videos and TigDog \cite{del2015articulated} that were excluded during training. 
In addition, we test our model's generalisation capacities using images depicting zebras \cite{zuffi20173d}. Note that we use the same trained model for all the cases;  surprisingly, the model still does well on the zebras, although the training set does not contain images of these animals. Fig. 3 shows the qualitative results of the 2D poses predicted by our trained model on different data. 


\begin{figure}[ht]
\centering
\includegraphics[width=\linewidth]{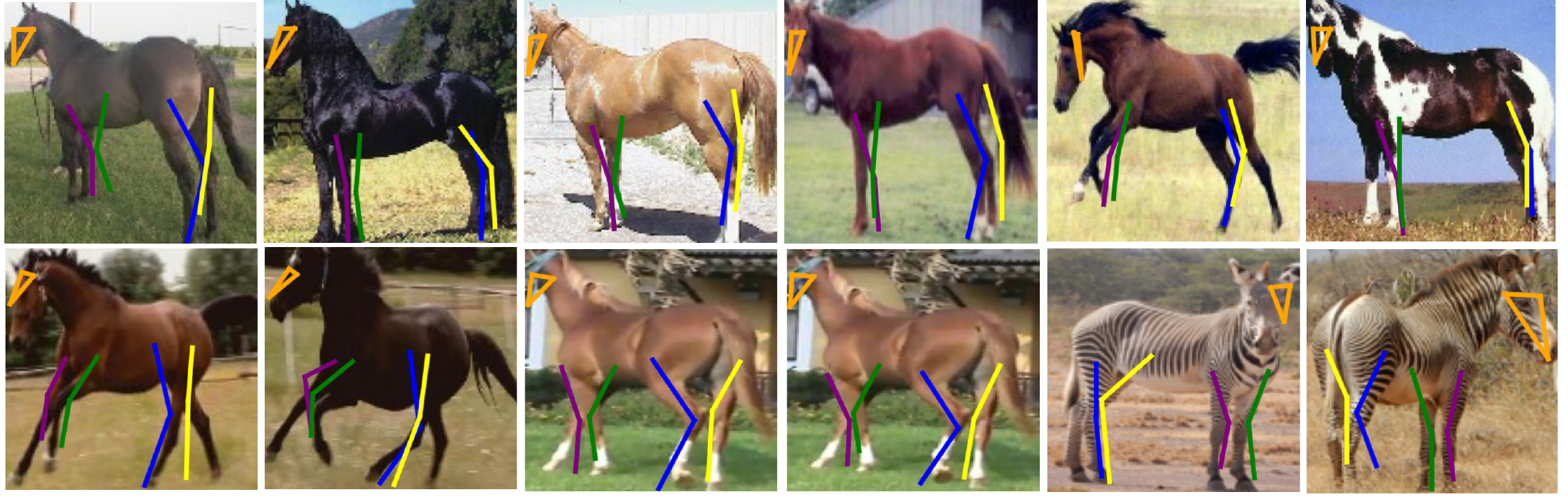}
\caption{\textbf{First row:} Predictions using images from \cite{borenstein2004combining}. \textbf{Second row:} Predictions using images from \cite{del2015articulated} and \cite{zuffi20173d}.}
\label{horse-results}
\end{figure}

\section{Discussion and Conclusion}\label{sec:discussion}
Supervised methods learn from annotated poses on the training data, which makes them dependent on the quality of those annotations. Although some joint positions are easy to annotate, others require domain specialists to locate them. Contrary to the supervised methods, our approach is not dependent on the quality of the annotations since it learns from skeleton images generated from synthetic poses. The method produces similar 2D poses to those obtained using DLC (Section C of Fig.\ref{all-results}), and its quantitative performance in terms of MPJPE against ground-truth annotations is not significantly different from DLC.

According to the plots in Fig.\ref{all-results}, despite some visible differences between our method and DLC for specific body parts, most graphs show smooth lines for our predictions. When comparing both methods against ground truth annotations, as expected, the overall performance of DLC is superior. This is probably due in part to the consistency of manual annotation of ground-truth joint locations used in both training and testing of DLC. Our experiment using synthetic images and a synthetic pose prior demonstrates that accurate predictions can be made by matching the pose prior and image domains. 

In conclusion, we successfully adapted a self-supervised 2D human pose estimation method to a different animal domain, replacing an empirical prior associated with actual 2D poses with a synthetic prior. We have demonstrated that the approach produces promising results compared to a state-of-the-art supervised approach in the mouse domain. An important motivation for our work has been to explore an approach that can be rapidly deployed to other animal domains without requiring extensive annotation of images. We demonstrate the latter qualitatively using a dataset of horse images. Finally, we plan to use our estimated 2D poses to measure gait on genetically modified mice with different levels of ALS disease. These measures could help to identify and classify patterns related to the development of the disease. 

\vspace{4ex}

\textbf{Ethics statement:} This study was approved by the University of Tasmania Animal Ethics Committee (permit number A17008) and designed in accordance with the Australian Code of Practice for the Care and Use of Animals for Scientific Purposes.

\vspace{2ex}
\textbf{Acknowledgments} Special thanks to Rebecca Stone and Mohammed Alghamdi from the School of Computing at the University of Leeds for great discussions and insightful feedback.

%
%
\bibliographystyle{samplepaper}
\bibliography{samplepaper}

\end{document}